# Multiscale Low-Frequency Memory Network for Improved Feature Extraction in Convolutional Neural Networks

**Fuzhi Wu**[1,2,3,4], **Jiasong Wu**[1,3,4,#], **Youyong Kong**[1,3,4], **Chunfeng Yang**[1,3,4], **Guanyu Yang**[1,3,4], **Huazhong Shu**[1,3,4,*], **Guy Carrault**[2,3,4], **Lotfi Senhadji**[2,3,4]

[1]Key Laboratory of New Generation Artificial Intelligence Technology and Its Interdisciplinary Applications (Southeast University)

[2]Laboratoire Traitement du Signal et de l'Image (Univ Rennes)

[3]Centre de Recherche en Information Biomédicale Sino-français (CRIBs)

[4]Jiangsu Provincial Joint International Research Laboratory of Medical Information Processing (Southeast University)

230198162@seu.edu.cn, jswu@seu.edu.cn, shu.list@seu.edu.cn,

## Abstract

Deep learning and Convolutional Neural Networks (CNNs) have driven major transformations in diverse research areas. However, their limitations in handling low-frequency information present obstacles in certain tasks like interpreting global structures or managing smooth transition images. Despite the promising performance of transformer structures in numerous tasks, their intricate optimization complexities highlight the persistent need for refined CNN enhancements using limited resources. Responding to these complexities, we introduce a novel framework, the Multiscale Low-Frequency Memory (MLFM) Network, with the goal to harness the full potential of CNNs while keeping their complexity unchanged. The MLFM efficiently preserves low-frequency information, enhancing performance in targeted computer vision tasks. Central to our MLFM is the Low-Frequency Memory Unit (LFMU), which stores various low-frequency data and forms a parallel channel to the core network. A key advantage of MLFM is its seamless compatibility with various prevalent networks, requiring no alterations to their original core structure. Testing on ImageNet demonstrated substantial accuracy improvements in multiple 2D CNNs, including ResNet, MobileNet, EfficientNet, and ConvNeXt. Furthermore, we showcase MLFM's versatility beyond traditional image classification by successfully integrating it into image-to-image translation tasks, specifically in semantic segmentation networks like FCN and U-Net. In conclusion, our work signifies a pivotal stride in the journey of optimizing the efficacy and efficiency of CNNs with limited resources. This research builds upon the existing CNN foundations and paves the way for future advancements in computer vision. Our codes are available at https://github.com/AlphaWuSeu/MLFM.

## Introduction

Deep learning (LeCun et al., 2015), particularly Convolutional Neural Networks (CNNs), has revolutionized diverse research fields and industrial applications, becoming a paramount tool for large-scale natural image processing. However, spurred by the transformative success of the Transformer model in NLP, researchers have attempted to merge its modules into vision CNN models (Carion et al., 2020, Bello et al., 2019). Vision Transformer (ViT) (Dosovitskiy et al., 2020) and its subsequent variants (Liu et al., 2021, Touvron et al., 2021, Yuan et al., 2021) have enjoyed remarkable success, even surpassing CNNs in supervised performance, challenging their long-standing predominance.

Yet, this evolution provokes questions about CNN's capabilities: has its "magic" waned? Are we tapping its full potential? It's time to re-examine this network and discover new capabilities. In traditional CNN structures, images are filtered through multiple convolution layers, eliciting high responses for salient patterns and discarding low responses via pooling layers. This design, however, struggles with low-frequency information. High responses often correspond to high-frequency details like textures, fine lines, and edges, leading to difficulties in eliciting high responses for low-frequency information. This bias results in faster discarding of low-frequency information as the network deepens, potentially hampering tasks involving global structures and large-scale patterns.

Recent research (Geirhos et al., 2018; Park et al., 2023) has shown a light on this deficiency: CNNs display a texture bias, often overlooking shape information. To rectify this, strategies have been employed, such as texture-independent datasets (Geirhos et al., 2018), discrete Fourier (Ryu et al., 2018) or wavelet pooling (Zhao and Snoek, 2021, Williams and Li, 2018, Li et al., 2020, Wang et al., 2021), skip connections and residual blocks, and designing networks with separate channels for low and high-frequency information (Monday et al., 2021; El Bouny et al., 2020).

*Corresponding author #Co-first author

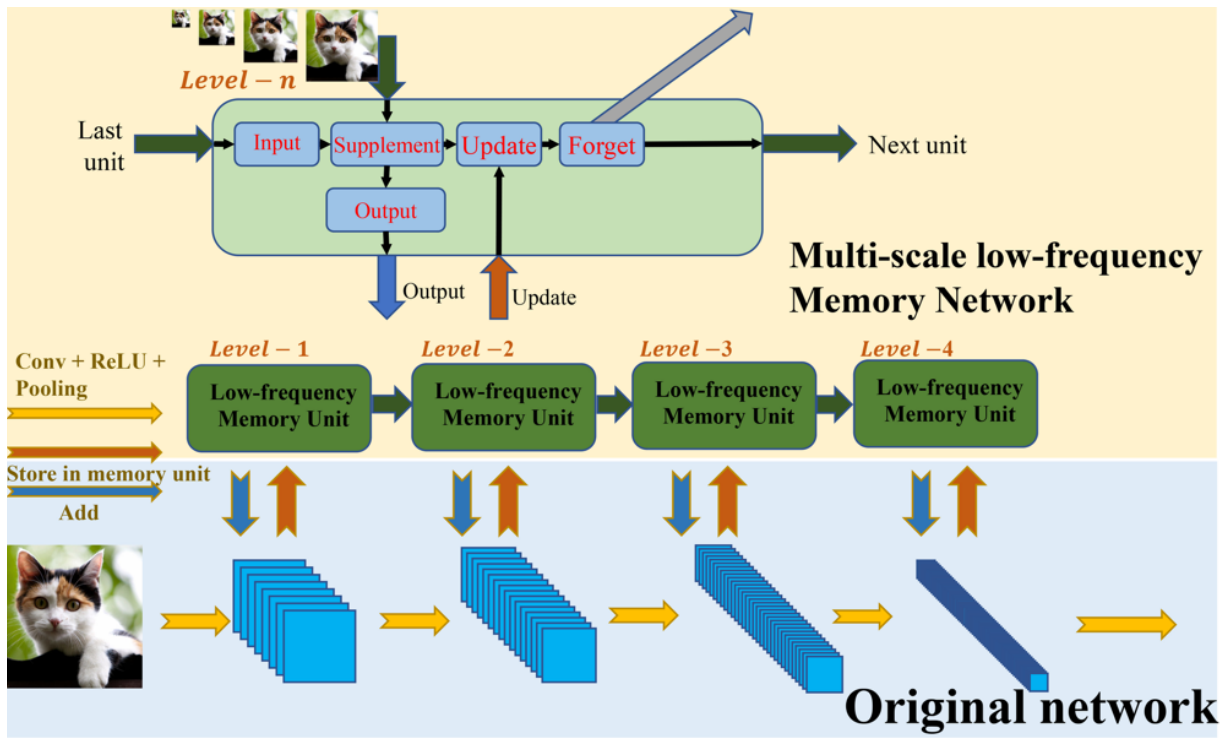

Figure 1: Our multiscale low-frequency memory (MLFM) framework. The bottom half presents the original network structure. We add a low-frequency information compensation branch to the network which will provide selectable low-frequency shape information for feature maps of different resolutions via multiple stacked LFMUs.

To efficiently retain and benefit from low-frequency information, we propose the Multiscale Low-Frequency Memory (MLFM) network as shown in Figure 1. The linchpin of MLFM is the Low-Frequency Memory Unit (LFMU), which creates a parallel channel to the core network. LFMUs store early features' low-frequency information, current features' low-frequency information, and multiscale low-frequency wavelet coefficients of original images. During forward propagation, LFMUs interact with features at each down-sampling step, ensuring the network learns and retains task-relevant low-frequency information. Our contributions can be summarized as follows:

- We present a multiscale low-frequency memory (MLFM) network, which could effectively provide low-frequency information storage to the network, reduce the loss of shape and color information and provide information interaction to the network at the key nodes.
- We design LFMUs by implanting five different forms of gates, in which multiscale wavelet transform is used for information extraction and the memory unit can raise the performance of CNNs ably.
- The proposed MLFM, which does not change the original core network structure, is a plug-and-play component that can be easily embedded in the most popular networks. The framework is validated with extensive experiments on ImageNet, and achieves considerable improvement in accuracy in many popular 2D CNNs. Its applicability is also established in image-to-image translation tasks, as evidenced by significant performance enhancements in semantic segmentation networks like FCN and U-Net.

## Related Work

**Wavelets and CNNs:** As a powerful time-frequency/time-scale analysis tool, not only the wavelet transform has gained great attention in the field of signal processing, but the combination of wavelet analysis with CNNs has also become a hot research topic in recent years. Many completely different combinations have been developed, which can be roughly divided into two main categories (Guo et al., 2022): preprocessing method and compact combination between the two (wavelets and CNNs).

- Preprocessing method: wavelet domain as a special feature space which can both show time and frequency information allows for better decomposition and reconstruction of the signal. Using the wavelet coefficients as the input feature vector to the CNNs rather than the spatial coefficients seems to be taken for granted (Li et al., 2021, Nneji et al., 2021). However, wavelet features emphasize the relevance of global information, while conventional convolution kernels in CNNs are more inclined to extract local chunked features, and it is difficult to exploit the full potential of a network by using wavelet coefficients as input.
- Compact combination between the two: integration of wavelets into CNNs as a processing module during forward propagation, has been a more popular practice in recent years. Li et al. (Li et al., 2020) hold the view that noise in the high-frequency components could be down-sampled into the following feature maps, and degrade the noise-robustness of the CNNs. Meanwhile, the basic object structures presented in the low-frequency component could be broken. Also, max pooling will encounter overfitting problems (Williams and Li, 2018). The traditional pooling layer has flaws that are difficult to address. Williams and Li (Williams and Li, 2018) put forth a suggestion that wavelet pooling can be an alternative to traditional neighborhood pooling which decomposes features into a second level and discards the first-level sub-bands to reduce feature dimensions. Many researchers have done more in-depth research on this aspect of work (Wolter and Garcke, 2021, Ferrà et al., 2018). Common in these works is the retention of the low-response value part of the convolutional layer in the learning process, rather than discarding it, but at the cost of sacrificing high-frequency edge and texture features containing more explicit semantic information. Completely replacing max-pooling with wavelet pooling is not ideal. Our framework offers an alternative low-frequency information storage unit without compromising the high-response edge information extracted by the CNN, effectively resolving the coexistence problem of low and high-frequency information.

**Multiscale CNNs:** Most existing CNN-based deep learning models experience spatial information loss and inadequate feature representation. This is attributed to their inability to capture multiscale background data and the semantic information loss during pooling (Elizar et al., 2022). Incorporating multiscale capabilities within the network can optimize feature capture at various scales, aligning better with the task at hand. Moreover, merging low-level and high-level features from different receptive fields can significantly enhance the models' performance. While various frameworks exist for integrating multiscale capabilities, they largely fall into two main categories (Elizar et al., 2022): multiscale feature learning and multiscale feature fusion.

- Multiscale feature learning: Classic work includes the multiscale convolutional networks proposed by Sermanet (Sermanet and LeCun, 2011), wherein a multi-stage architecture incorporates both high-resolution primary features and secondary low-resolution features into the classifier. Stofa et al. (Stofa et al., 2022) introduces a unique multiscale deep learning technique to recognize emotions from micro-expressions, by employing optimized spatial pyramid pooling and atrous spatial pyramid pooling methods, and integrating four new multi-layer convolutional network architectures that yield improved recognition performance.
- Multiscale feature fusion: Combining important information from multiple images into a single image, namely image-level fusion, can produce a more useful and comprehensive representation compared to the original input. In (Peng et al., 2019), the authors present a multi-sensor fusion SLAM and a swift dense 3D reconstruction pipeline, which offer roughly registered image pairs to a Deep Deconvolutional Network (DN) for precise pixel-wise change detection. Another conventional method, known as feature-level fusion, merges high-resolution features lacking rich semantic details with low-resolution features that are semantically enriched. In (Lin et al., 2017) Feature Pyramid Network (FPN) stands as a commonly employed strategy in object detection. Their approach constructs cost-effective feature pyramids and develops a top-down structure with lateral connections, facilitating the creation of high-level semantic feature maps across all scales. Our strategy combines image-level and feature-level multiscale fusion by integrating different scale information of the original image with compressed images obtained through wavelet multi-level decomposition in the supplementary gate. Additionally, storing wavelet coefficients of the bottom CNN layer in the memory unit and providing them to the top convolution layer enriches the high-level features with semantic, shape, and location information from the bottom layer simultaneously.

## Method

Our method aims at designing LFMU and constructing a MLFM network. We first introduce the main motivation of our method and the structure is then described in detail.

### Multiscale Low-Frequency Memory (MLFM)

**Motivation.** In contemporary CNNs, convolution and pooling layers are key components. Typically, an input image is first sequentially filtered by multiple convolutional layers, leading to high responses in distinctive features, and then the lower responses are discarded by pooling. This process usually yields high responses in regions of the image where intensity changes dramatically, embodying information about textures, fine lines, and edges, all high-frequency components of the image. This is illustrated in Figure 2, featuring feature maps from CNN's mid-layer using the Resnet34 (He et al., 2016) model in Pytorch (Paszke et al., 2019), with randomly initialized parameters.

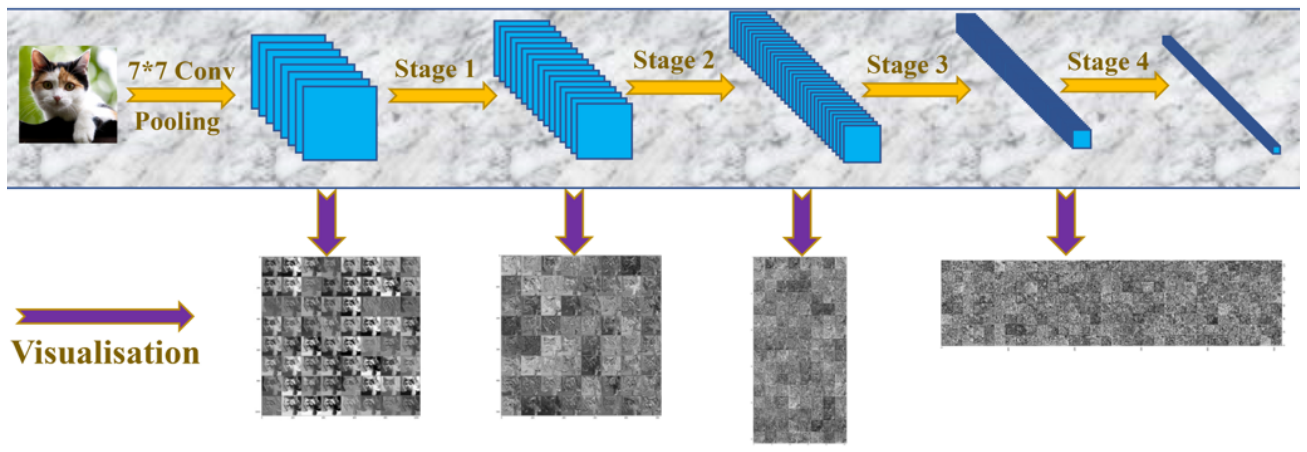

Figure 2: The visualization of a residual network. We can see that as the network deepens, it tends to extract more edge and texture features and the low-frequency components are gradually lost.

In essence, the Conv-Pool structure tends to retain image edges and texture features while discarding low-frequency information. For a quantitative evaluation, we apply the Average Structural Similarity Measure (SSM) (Wang et al., 2004), mainly utilized for comparing the similarity of two images of the same size. We find that CNNs inherently favor retaining high-frequency details, losing low-frequency components in the forward process. While high-frequency edge information is crucial for object classification, the preservation of low-frequency components that carry substantial image information is also necessary. Hence, our MLFM design seeks to preserve these harder-to-retain low-frequency components and to provide them to subsequent layers.

Further, as deep neural networks evolve from Vgg19(Simonyan and Zisserman, 2014) to ResNet34, and from ResNet101 to DenseNet121(Huang et al., 2017), network depth increase implies a higher ability to extract high-frequency information. Conversely, the original image's low-frequency information undergoes an irreversible loss. The low-frequency structure of the original image has a notably low similarity with the deep layer features of the network, necessitating additional retention of the original images' low-frequency information in our multiscale memory unit. Consequently, our MLFM framework comprises multiple LFMUs, each storing the low-frequency information of the previous unit, the current layer, and the original image.

**The design of MLFM framework.** Our MLFM framework is depicted in Figure 1. CNNs consist of two distinguished components: the feature extraction layer and the down-sampling layer. We use the latter to establish nodes for creating multiscale LFMUs. After down-sampling features, an information exchange channel is set up between the core network and the parallel low-frequency memory branch. Here, low-frequency data is transferred from the memory to the core network and vice versa. Crucially, the unit preserves low-frequency information across different scales, correlating with the current layer's resolution in the core network.

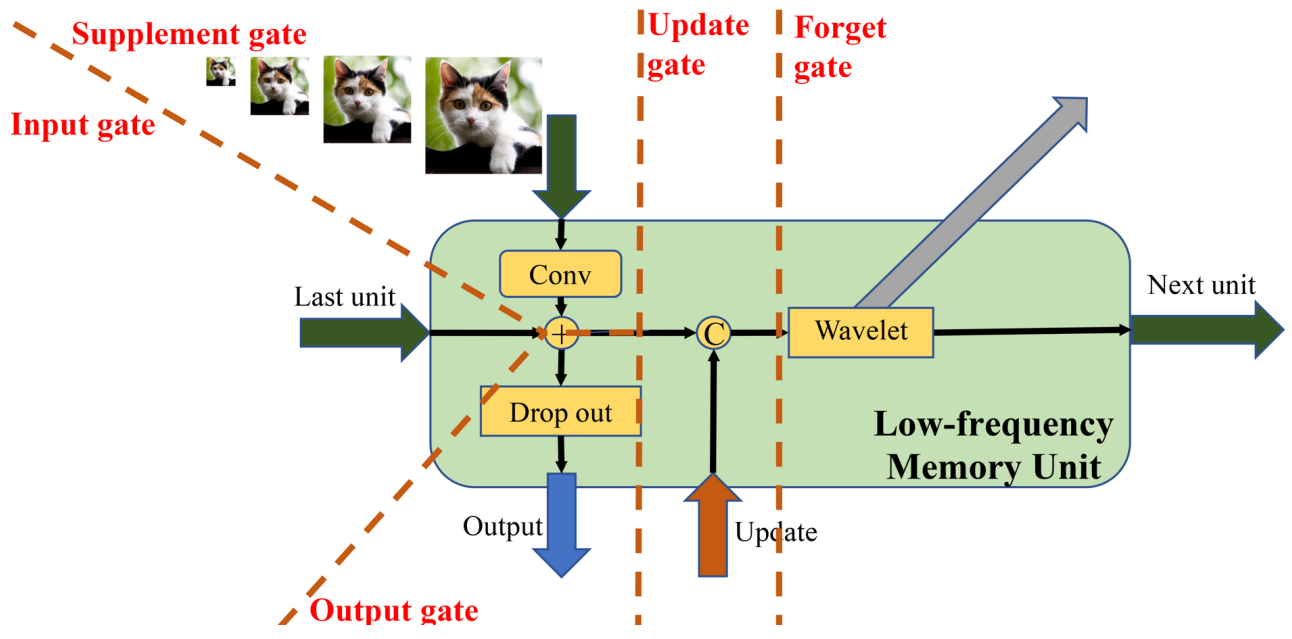

Figure 3: The design of low-frequency memory unit (LFMU). The whole memory unit consists of five main gates: input gate, output gate, supplement gate, update gate, and forget gate.

## Low-Frequency Memory Unit

Our LFMU contains five basic gate components as shown in Figure 3.

**Input gate:** To minimize computation, only the basic input layer is included in the gate to provide low-frequency information from the previous memory unit directly to the current unit.

**Update gate:** To provide low-frequency information to subsequent memory units, we design an update gate that combines the newly uploaded features with the memory's low-frequency information using the "cat" operation.

**Supplement gate:** To retain low-frequency features from the original image, we provide the memory unit with multiscale low-frequency information of original images. To minimize computation, this gate consists of only one essential convolutional layer for dimensionality consistency, directly adding the multiscale wavelet low-frequency coefficients to the memory unit.

**Forget gate:** As the network deepens, the cumulated information in memory units increases the memory burden. Given that deep networks don't necessitate high-resolution features, forgetting some information becomes essential. In our forget gate design, we discard high-frequency information—likely learned in the core network—via wavelet decomposition. We retain low-frequency features—challenging to learn in the core network—for future memory units.

**Output gate:** During the information exchange between the memory unit and the core network, the memory unit compensates the core network with stored low-frequency information. We avoid designing extra operations, instead adding only a basic dropout layer to prevent overfitting.

# Experiment

## Implementation Details

**ImageNet100:** We evaluate MLFM framework using various configurations, benchmarking against seven widely used CNNs (ResNet (He et al., 2016), MobileNet (Sandler et al., 2018), SENet (Hu et al., 2018), RegNet (Radosavovic et al., 2020), EfficientNet (Tan and Le, 2019), ConvNeXt (Liu et al., 2022), InceptionNeXt (Yu et al., 2023)) on a randomly selected 100-class of ImageNet (Deng et al., 2009). We utilize 130,000 randomly cropped (256*256) training images, and 5,000 testing images. Training employs an SGD optimizer (momentum 0.9, weight decay 1e-4), initial learning rate of 0.1, and batch size of 128. We adopt a MultiStepLR learning rate scheduler, limiting training to a maximum of 90 epochs.

**ImageNet1000:** To evaluate on large-scale dataset, we test on ImageNet-1K. The completed dataset of ImageNet is used and the parameter configurations are similar to ImageNet100.

**Cityscapes:** We test MLFM on Cityscapes dataset (Cordts et al., 2016) for image segmentation evaluation. Input images are cropped to 768*768 and randomly flipped. Cityscapes, comprising stereo video sequences from 50 cities' street scenes, is designed for semantic and instance tasks in autonomous driving. It contains over 5,000 finely annotated frames and 20,000 weakly annotated frames, spanning 30 diverse urban object and structure categories. We utilize the 5,000 finely annotated images and test on 2,975 valid images. During training, we use an SGD optimizer with 0.9 momentum, 1e-2 weight decay, an initial learning rate of 0.01, and a per-GPU batch size of 4. Our loss function is ProbOhemCrossEntropy2d and training is capped at 200 epochs.

All models are implemented in PyTorch and with one Nvidia RTX 4090 GPU.

## Ablation Study

Our ablation studies focus on three aspects. Firstly, we experiment by placing multiple LFMUs at different locations. Secondly, we contrast original LFMU with versions missing certain components. Finally, we compare the effects of various wavelet basis functions. In this section, we conduct experiments on ImageNet100 using ResNet18 as the backbone.

**Different positions of LFMUs.** Our MLFM situates LFMUs at several nodes within the core network, specifically at nodes corresponding to features after each down-sampling layer. For a typical CNN like ResNet, this involves 4-5 down-sampling points, as depicted in Figure 3. We experiment with different node combinations and position our LFMUs accordingly, with results shown in Table 1. Here, 'baseline' refers to the original ResNet18 performance on ImageNet100, devoid of LFMUs. 'Li-Lj' signifies adding LFMU units at each layer from Li to Lj, where 'i=j' indicates adding LFMUs only at that layer. Evidently, LFMUs enhance network performance at every location, demonstrating the significance of low-frequency information compensation. Placement of LFMUs in L1-L5 yields the best results, boosting accuracy on ImageNet by 3.36% relative to the original network.

**Supplement gate in LFMU.** In MLFM, each memory unit is designed with essential input and output gates. Additionally, we incorporate low-frequency information from the original image at different scales as a supplementary gate within the memory unit. Here, we compare the performance of LFMU with this supplementary gate against an original LFMU lacking this gate. Experimental results are showcased in Table 1. They substantiate the efficacy of our supplementary module, demonstrating a consistent enhancement in LFMU's performance.

**Wavelet transform in LFMU.** In our LFMU design, we leverage wavelet transform as a down-sampling tool. The forget gate utilizes wavelet transform to retain the low-frequency components in the memory unit, while high-frequency components are discarded. Similarly, the supplement gate derives multiscale information of the original image via various levels of wavelet decomposition. The backbone network's Conv-Pooling structure is adept at learning the high-frequency information necessary for the target task, yet it struggles to retain certain low-frequency information. Hence, we use wavelet transform to preserve low-frequency components at different scales within the memory units. Table 2 demonstrates the impact of substituting wavelet decomposition by max-pooling and average pooling in both the forget and supplement gates. These experiments reveal that wavelet sampling leads to the highest classification recognition rate.

**Different basis functions in LFMU.** This section explores the impact of different wavelet basis functions on results. We test a variety of wavelet basis functions: Haar, Biorthogonal, Daubechies, Symlets, Coiflet, and Dmeyer. Table 3 displays the Top-1 and Top-5 accuracy of MLFM_ResNet18 on the ImageNet100 validation set, with different wavelet types designated by various codes. The length of wavelet filters increases with orders. Although Haar wavelets are symmetric, Daubechies and Symlets are not. The results reveal that different wavelet basis functions significantly affect network performance. Biorthogonal wavelet functions are relatively unaffected by different orders, while performance decreases with increasing order for Daubechies wavelets. Symlets wavelet function performance varies with a sample variance of 0.2547 across different orders, and a clearer performance decline is observed for the Coif wavelet function as the order increases. Overall, the Haar wavelet achieves the highest Top-1 accuracy.

| Networks | With supplement gate | No supplement gate |
|---|---|---|
| Baseline | 77.86% | - |
| L1-L1 | 78.52% | - |
| L1-L2 | 78.94% | 78.40% |
| L1-L3 | 78.66% | 78.02% |
| L1-L4 | 80.58% | 79.88% |
| **L1-L5** | **81.22%** | 79.36% |
| L2-L2 | 78.16% | - |
| L2-L3 | 79.36% | 78.34% |
| L2-L4 | 79.98% | 78.52% |
| L2-L5 | 80.12% | 78.96% |
| L3-L3 | 78.36% | - |
| L3-L4 | 79.52% | 78.11% |
| L3-L5 | 79.14% | 78.16% |
| L4-L4 | 78.84% | - |
| L4-L5 | 78.64% | 77.44% |
| L5-L5 | 79.86% | - |

Table 1: Comparison of the effects of different positions of LFMUs with and without supplement gate.

| Networks | Top-1 Acc | Top-5 Acc |
|---|---|---|
| Original network | 77.86% | 93.74% |
| MLFM_Maxpooling | 77.64% | 93.86% |
| MLFM_Averagepooling | 77.68% | 93.25% |
| MLFM_Wavelet | **81.22%** | **95.08%** |

Table 2: Comparison of the effects of supplement gate in LFMUs.

| Wavelet basis | | Top-1 Acc | Wavelet basis | | Top-1 Acc |
|---|---|---|---|---|---|
| Haar | | **81.22%** | Dmeyer | | 79.86% |
| Biorthogonal | bior1.1 | 80.48% | Daubechies | db1 | 79.80% |
| | bior1.3 | 80.58% | | db4 | 79.76% |
| | bior2.2 | 80.40% | | db8 | 79.60% |
| | bior3.3 | 80.40% | | db16 | 77.54% |
| Symlets | sym2 | 79.66% | Coif | coif1 | 80.84% |
| | sym4 | 80.58% | | coif2 | 80.36% |
| | sym8 | 80.96% | | coif4 | 79.10% |
| | sym20 | 79.52% | | coif8 | 64.74% |

Table 3: Comparison of the effects of different wavelet basis functions in LFMUs.

## MLFM vs. Traditional Frameworks: Performance and Complexity Analysis

In this section, we assess our MLFM 's versatility by applying it to various popular networks using the ImageNet100. We incorporated our LFMU into ResNet, a notable network, and compared the parameter count and error rate as depicted in Figure 5. The original ResNet is denoted by the red line, and the blue line represents our MLFM ResNet. The vertical axis corresponds to the error rate, while the horizontal axes represent the parameters. Our MLFM framework significantly improves the ResNet architecture, consistently reducing the error rate compared to the original network.

In the realm of lightweight deep learning models, we present a comparative analysis of our network's results on MobileNetV2, illustrated in Figure 5. Figures 6, and 7 highlight the performance of our LFMU unit when integrated into more recent, popular, and efficient frameworks (EfficientNet, SeNet). The ConvNeXt network, launched in 2022, also shows promising results with our network, as indicated in Figure 8. Additionally, in InceptionNeXt, released in 2023, a comparative performance evaluation is demonstrated in Table 4.

Our MLFM network offers a plug-and-play enhancement to most mainstream CNN networks, bolstering their effectiveness without necessitating changes to the original network architecture. This allows CNNs to display superior performance trends.

## Effectiveness on Large-Scale Data

After demonstrating our framework's versatility across different network structures, we aim to further validate its efficacy for large-scale, comprehensive datasets. Implementing an LFMU at levels L1-L5 on the original network, we select "Haar" as our wavelet basis function. We conduct experiments on ResNet18 and ResNet34, with results presented in Table 5. These results affirm that our model maintains high performance when applied to extensive, large-scale datasets.

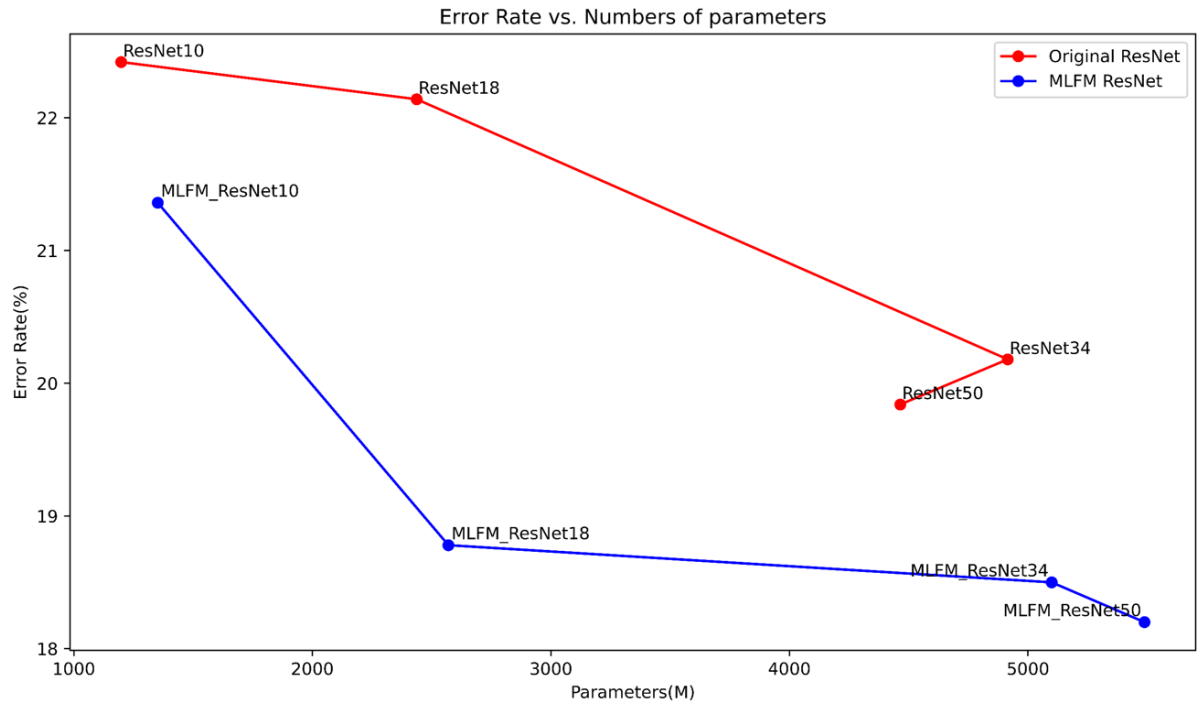

Figure 4: Performance of MLFM in residual networks. The numbers following the networks represent layers of different ResNet. Error Rate vs. Number of parameters in original ResNet and MLFM.

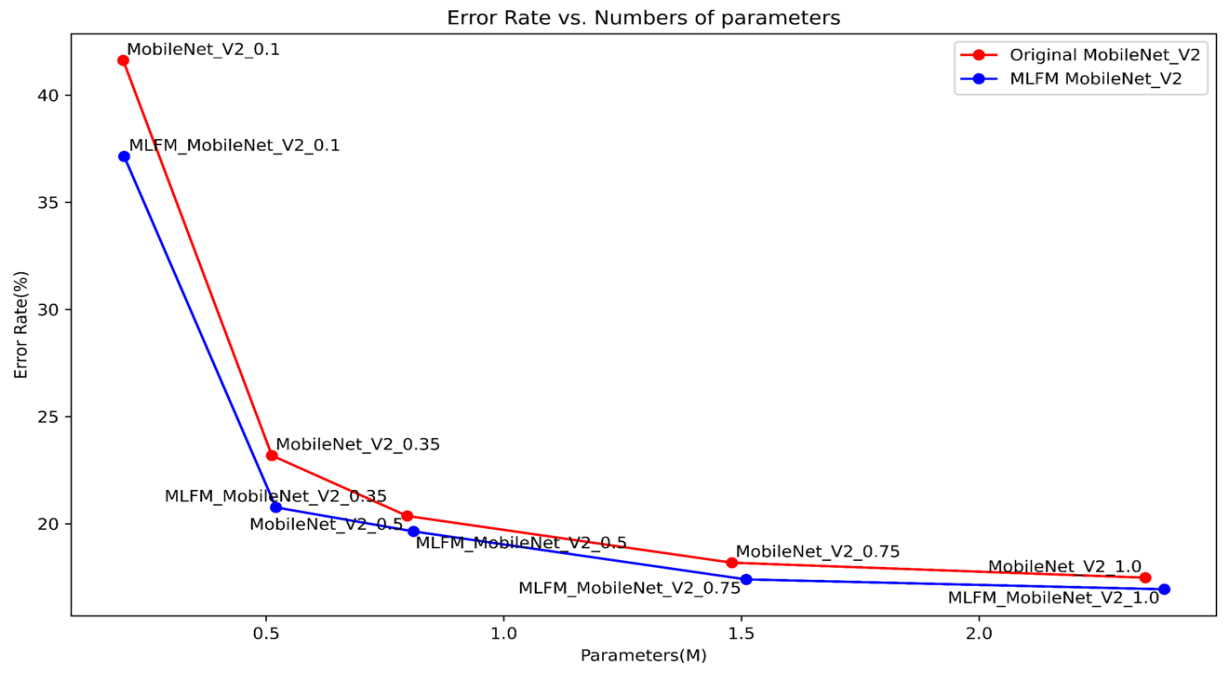

Figure 5: Performance of MLFM in MobileNetV2.

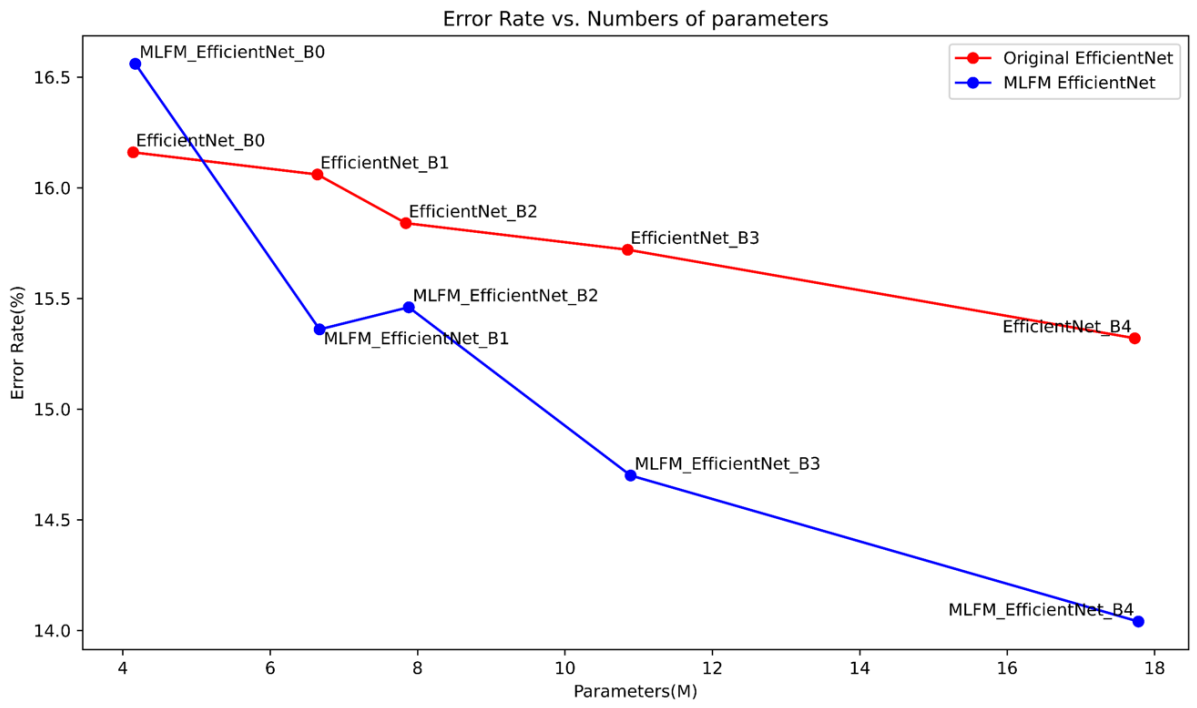

Figure 6: Performance of MLFM in EfficientNet.

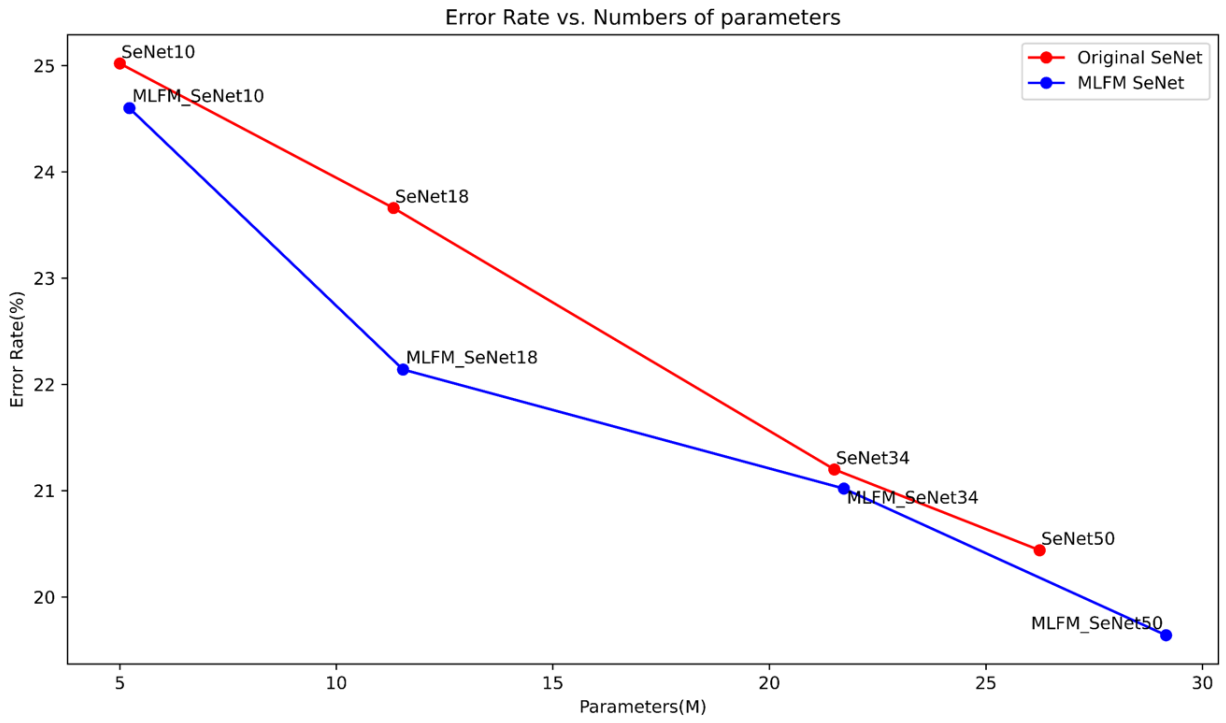

Figure 7: Performance of MLFM in SeNet.

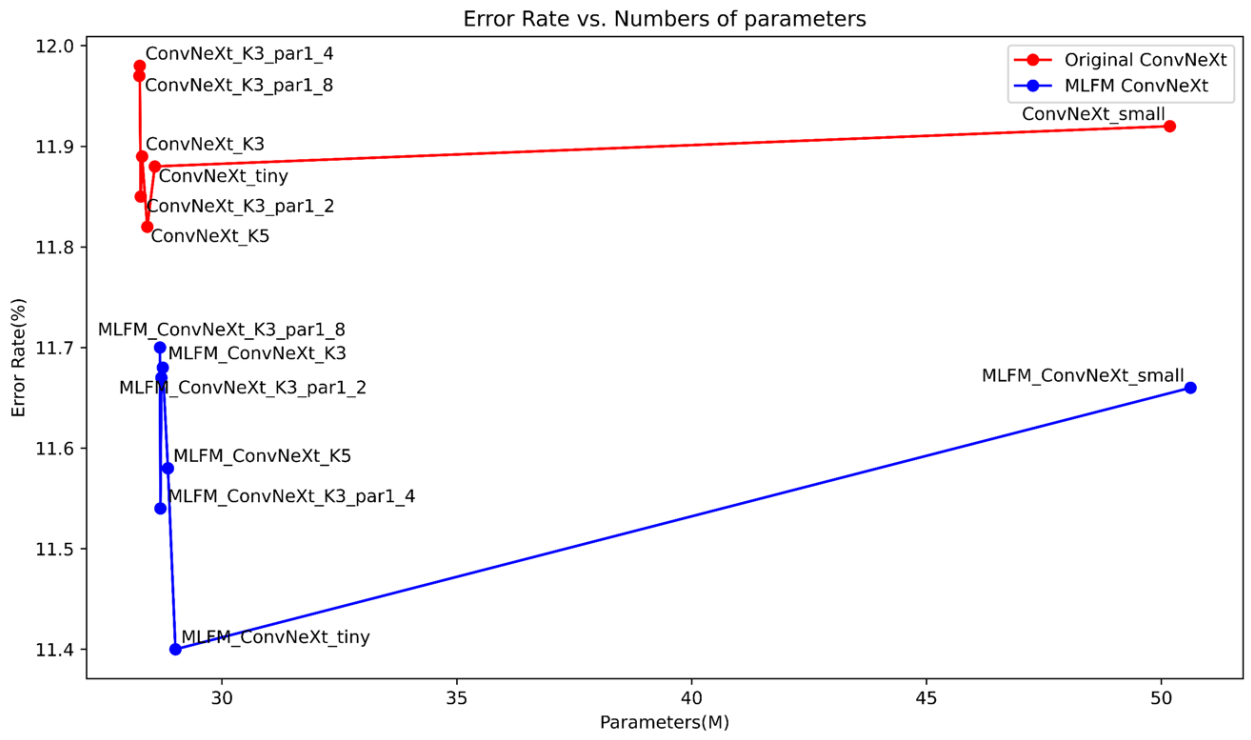

Figure 8: Performance of MLFM in ConvNeXt.

| Network | Parameters | MACs | Accuracy |
|---|---|---|---|
| InceptionNeXt | 28.05M | 5.49G | 87.58% |
| MLFM_ InceptionNeXt | 28.50M | 5.76G | 88.06% |

Table 4: Performance of MLFM in InceptionNeXt.

| Network | Parameters | MACs | Accuracy |
|---|---|---|---|
| ResNet18 | 11.69M | 2.38G | 69.79% |
| MLFM_ ResNet18 | 12.09M | 2.57G | 71.32% |
| ResNet34 | 21.80M | 4.80G | 73.09% |
| MLFM_ ResNet34 | 22.80M | 5.06G | 74.66% |

Table 5: Comparison between MLFM and common CNNs on ImageNet1000 dataset.

## Image Segmentation

Our key contributions include the LFMU module and MLFM framework. LFMU is a plug-and-play module proficient in retaining low-frequency information within networks, enabling easy adaptation of the MLFM for image-to-image translation tasks. We demonstrate MLFM's potential in semantic image segmentation tasks.

To evaluate the potential of our module, we perform experiments on two notable networks, FCN_ResNet and UNet on Cityscapes dataset. These semantic segmentation networks adopt an encoder-decoder structure. Our experiment initially deploys the LFMU only in the encoder, then in both encoder and decoder sections. We use FCN_ResNet with ResNet18 as the encoder's backbone and a max-unpooling layer in the decoder for restoring the original resolution, as shown in Figure 9(a). We slightly adapt MLFM for segmentation, with the LFMU retaining low-frequency information in the encoder for a consistent circulation of such features, and dispatching discarded high-frequency information to the decoder, illustrated in Figure 9(b).

We compare the original segmentation networks and MLFM versions on Cityscapes dataset. Table 6 shows the results of the original network, the network with LFMU only in the encoder (MLFM_encoder), and the network with LFMU in both encoder and decoder (MLFM_encoder-decoder). Including the LFMU in the encoder greatly enhances the performance of the semantic segmentation network, maintaining effectiveness when extended to the decoder.
In Figure 10, we provide a detailed analysis and visual examples for each category of FCN_ResNet, MLFM_ResNet_encoder, and MLFM_encoder-decoder. Images cover "buildings", "trees", "fence", "pole", "sky", "bus", and "cars" regions, along with segmentation results using three networks. In Example I, FCN_ResNet struggles to differentiate a bus from cars and trucks. Once we integrate the LFMU into the encoder, thus leveraging low-frequency shape information, the network starts correctly identifying the bus, with further improvement when the LFMU is added to the decoder. Similarly, in Example II's top right corner, FCN_ResNet blurs the "pole", "sky", and "buildings", whereas MLFM_FCN clearly distinguishes these elements.

| Network | OA | MIoU | Precision | Recall |
|---|---|---|---|---|
| UNet | 95.81% | 66.25% | 74.18% | 83.47% |
| UNet_MLFM_encoder | 95.84% | 67.29% | 75.45% | 83.59% |
| UNet_encoder_decoder | 95.87% | 68.32% | 76.30% | 84.71% |
| FCN_ResNet | 95.09% | 64.32% | 73.80% | 80.72% |
| FCN_MLFM_encoder | 95.33% | 65.27% | 74.07% | 82.52% |
| FCN_encoder_decoder | 95.42% | 67.43% | 76.18% | 83.27% |

Table 6. Comparison between MLFM and semantic image segmentation CNNs on Cityscapes dataset.

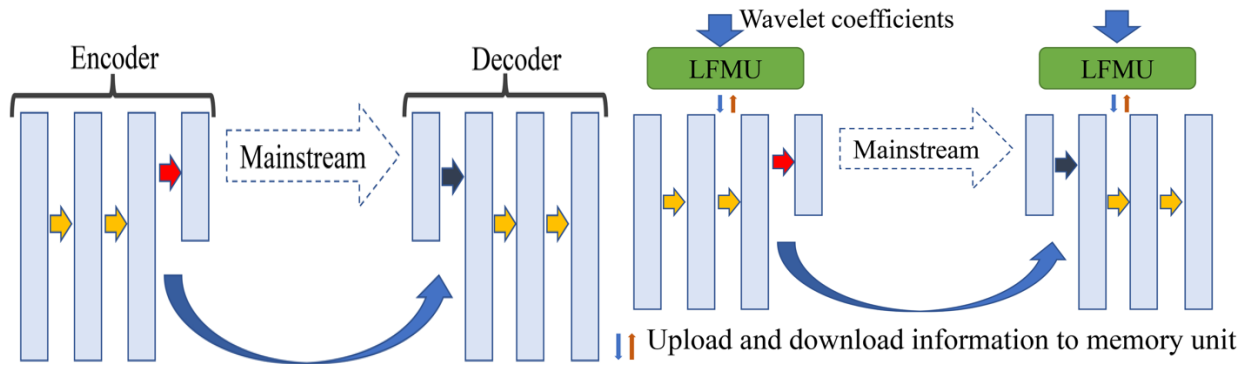

Figure 9: Basic down-sampling and up-sampling module of FCN and MLFM_FCN. (a) FCN. (b) MLFM FCN.

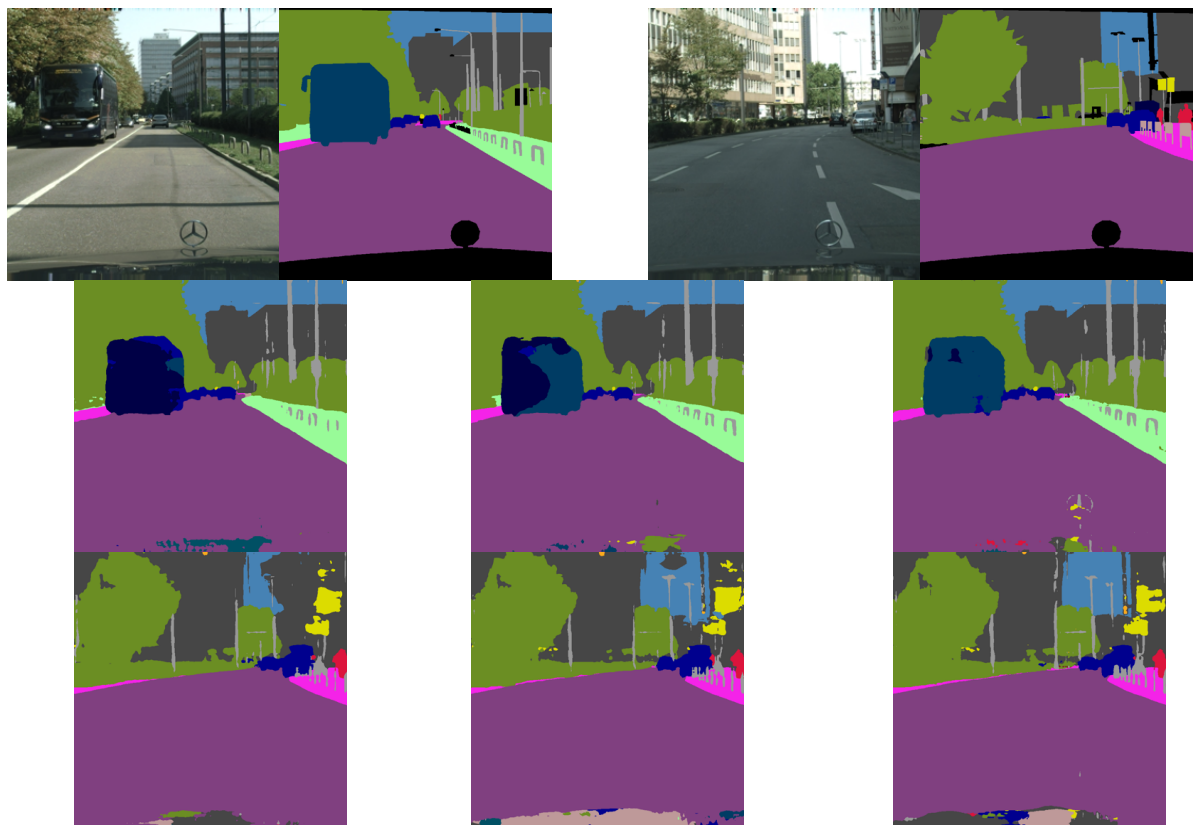

a) FCN_ResNet. b)MLFM_encoder. c)MLFM_encoder_decoder.

Figure 10: Comparison between FCN_ResNet and FCN_ResNet_MLFM segmentation.

## Discussion

This research addresses the limitations of CNNs in capturing low-frequency information, emphasizing their importance for network robustness and versatility. The proposed Multiscale Low-Frequency Memory (MLFM) network, centered around the Low-Frequency Memory Unit (LFMU), effectively preserves and manages low-frequency information.
LFMU's effectiveness lies in its intricate design, featuring gates such as Input, Update, Supplement, Output, and Forget, enhancing dynamics, adaptability, and integration of original image information. This streamlined MLFM framework enables seamless integration into various existing networks without altering their core structures. Experimental validation demonstrates significant accuracy improvements across a range of CNNs.

Future research should focus on optimizing MLFM for diverse applications and data types, refining LFMU's design, and understanding the contributions of different gates to overall performance.

## Acknowledgments

This work was supported in part by the National Key Research and Development Program of China (Nos. 2022YFE0116700, 2021ZD0113202), and in part by the National Natural Science Foundation of China under Grants 62171125, 61876037, and in part by the innovation project of Jiangsu Province under grants BZ2023042, BY2022564.